\documentclass[conference]{IEEEtran}
\ifCLASSINFOpdf
\usepackage{amsmath}

\DeclareMathOperator*{\argmin}{arg\,min}
\usepackage{amssymb}
\usepackage{graphicx,subfigure}
\usepackage{algorithm2e}
\usepackage[backend=bibtex,style=numeric-comp,sorting=none]{biblatex}
\usepackage{xcolor}
\usepackage{gensymb}
\usepackage{float}

\bibliography{b} 
\graphicspath{{figures/}}
\begin{document}

\title{A Comparative Study of Coarse to Dense 3D Indoor Scene Registration Algorithms }

\author{
	\IEEEauthorblockN{Abdenour Amamra and Khalid Boumaza}
	\IEEEauthorblockA{Ecole Militaire Polytechnique, Bordj El-Bahri BP 17, Algiers, Algeria}
	\IEEEauthorblockA{Email: amamra.abdenour@gmail.com}	
}

\maketitle

\begin{abstract}
		3D alignment has become a very important part of 3D scanning technology. For instance, we can divide the alignment process into four steps: key point detection, key point description, initial pose estimation, and alignment refinement. Researchers have contributed several approaches to the literature for each step, which suggests a natural need for a comparative study for an educated more appropriate choice. In this work, we propose a description and an evaluation of the different methods used for 3D registration with special focus on RGB-D data to find the best combinations that permit a complete and more accurate 3D reconstruction of indoor scenes with cheap depth cameras.
\end{abstract}
\IEEEpeerreviewmaketitle
	
\section{Introduction}
Interest in the digitization of the real world has increased over the last few years. One way to achieve this operation can be through the utilization of RGB-D data and appropriate geometric representations. However, no matter accurate is a sensor, they mostly have a limited field of view. Holistic panoramic cameras suffer from serious distortions, thus rendering the 3D acquisition process prone to severe errors and discrepancies over time \cite{li2018automatic}. As a result, we need to gain several captures at different viewpoints with a simple camera. The problem is that we get different views of an object without knowing where they were captured, resulting in unconnected pieces of 3D points that need to be merged in the same reference frame to reconstruct the object of interest. 

Aligning images into a single 3D model is called \emph{3D registration} or \emph{alignment}. For instance, 3D alignment is a fundamental problem in various domains such as archaeological site reconstruction, medical imaging, shape recovery \cite{amamra2016recursive}. Specific problems include the alignment of temporal 3D images for lesion segmentation, structure modeling, and 3D object reconstruction. More recently, interest in virtual and augmented reality applications has grown up, where 3D alignment algorithms are playing a crucial role, particularly with the rise of 3D virtual reality gadgets such as Microsoft HoloLens and Facebook's Oculus Rift.

The reconstruction of 3D scenes from images taken at different parts of the same scene has become possible with 3D alignment. However, the alignment of the views requires a division of the whole processing chain into several steps. Several methods, from computer vision literature, have been proposed so far for each of the aforementioned steps. The choice of using a method in a step depends on the nature of the scene and the technology behind the sensing device. For this reason, we set as an objective in the present work to perform a comparative study between the existing methods by testing the different 3D image alignment strategies. The camera we focused on for data acquisition is the Kinect V2. The latter captures 3D images in the form of colored 3D point clouds. 

The remainder of the paper is divided into three main sections, where we follow the logical and chronological sequence of our processing steps. In Section 2, we present the state-of-the-art regarding the different methods involved in 3D alignment. In Section 3, a series of tests is carried out, then a comparative study is achieved in order to determine the appropriate method for each task. Finally, we conclude by summarizing the obtained results and propose some potential future works.
\section{Background}

\subsection{Harris3D}
Harris \cite{sipiran2011harris} is a key point detector based on points and edges that are characterized by high intensity change in their horizontal and vertical neighborhood. In the 3D case, one can replace image gradients by surface normals for the same responses. To find the key points, we use a Hessian matrix of the intensity $C$ around a given 3D point. Such a matrix is smoothed by a Gaussian kernel $w_G(\sigma) $. That is $C_{Harris} = w_G(\sigma) * C $, such that $\sigma $ is the standard deviation of the filter and $*$ denotes the convolution operator. A measure of response at each key point  $(x, y, z)$ is then defined by :
\begin{equation}
	r(x, y, z) = det(C_{Hrs}(x, y, z))- k (tr(C_{Hrs}(x, y, z)))^2
\end{equation}
where $k$ is a positive real valued parameter. This parameter serves roughly as a lower bound for the ratio between the magnitude of the weaker edge and that of the stronger one.

\subsection{SIFT3D}

The Scale Invariant Feature Transform (SIFT) key points \cite{flitton2010object} are encoded by vectors that measure the local neighborhood. The original algorithm for 3D point clouds uses a 3D version of the Hessian to select the key points. A density function $f(x, y, z)$ is approximated by sampling the data regularly in space. A scale space is	built over the density function, and a local maxima search of the Hessian determinant  is performed.
The input cloud, $I(x, y, z)$ is convolved with a number of	Gaussian filters whose standard deviations ${\sigma_1,\sigma_2, ... }$ differ by a fixed scale factor. That is, $\sigma_{j+1} = k\sigma_j$, where $k$ is a
constant scalar that should be set to $\sqrt{2}$. The convolutions yield smoothed images denoted by :
\begin{equation}
	G(x, y, z, \sigma_j), i = 1, ... , n
\end{equation}
The adjacent smoothed images are then subtracted to yield a small number (3 or 4) of Difference-of-Gaussian (DoG)
clouds by:
\begin{equation}
	D(x, y, z, \sigma_j) = G(x, y, z, \sigma_{j+1}) - G(x, y, z, \sigma_{j})
\end{equation}

\subsection{ISS3D} 
Intrinsic Shape Signatures (ISS) \cite{salti2011performance} is a method relying on region-wise quality measurements. This method uses the magnitude of the smallest eigenvalue to include only points with large variations along each principal direction and the ratio between two successive eigenvalues  for excluding points having similar spread along principal directions. The intrinsic frame is a characteristic of the local object shape and independent of viewpoint.  

\subsection{SUSAN} 
The Smallest Univalue Segment Assimilating Nucleus (SUSAN) corner detector has been introduced in \cite{filipe2014comparative}. Rather than testing local gradients, a morphological approach is used; where for each pixel in the image, we consider a circular neighborhood of a fixed radius. The central pixel is referred to as the nucleus. Then, all other pixels within this circular neighborhood are partitioned into two categories: \emph{similarity} or \emph{differentiation}, depending on whether they have similar intensity values as the nucleus. This way, each point cloud has an associated local area of similar brightness, whose relative size contains important information about the structure of the cloud at that point. 

\subsection{SHOT} 
An evaluation of most existing key point descriptors has shown that the de definition of a unique local frame (a coordinate system) at each key point is the most important problem in the description of the key points \cite{aldoma2012tutorial}. In addition, the authors proposed a new descriptor based on a local reference frame (RF). The latter was called the Signature of Histograms of Orientations (SHOT). SHOT belongs to the signature descriptors of the histogram family \cite{aldoma2012tutorial}. It promotes the efficiency of calculation and the distinctive character and robustness against noise. 

\subsection{FPFH}
FPFHs (Fast Point Feature Histograms) are informative local features that represent the surface model properties underlying a point $p$ \cite{aldoma2012tutorial}. Their calculation is based on the combination of certain geometrical relations between the $k$ nearest neighbors of $p$. They incorporate the coordinates $(x, y, z)$ of the 3D point and the estimated surface normal $(n_x, n_y, n_z)$.
\subsection{Iterative Closest Point}
\subsubsection{Point-to-point variant}
The point-to-point ICP algorithm was originally proposed in \cite{aldoma2012tutorial} and obtains point mappings by looking for the target closest point $q_i$ to a point $p_j$ in the source point cloud. The closest neighbor correspondence is measured in terms of the Euclidean distance metric:
\begin{equation}\label{8}
	\hat{i}=\argmin_i ||p_i-q_j||^2
\end{equation}
Where $i  \in [0, 1, ..., N]$, and $N$ is the number of points in the target point cloud. The rotation parameters $ R $ and translation $ t $ are estimated by minimizing the distance between the corresponding pairs:
\begin {equation}\label{9}
\hat{R}, \hat{t} = \argmin_{R, t} \sum_{i = 1}^{N} ||(R \times p_i + t) - q_j||^2
\end {equation}
ICP resolves iteratively \ref{8} and \ref{9} to improve estimates of previous iterations. 

\subsubsection{Point-to-plane variant}
Because of the simplistic definition of point mappings, the point-to-point ICP algorithm is rather sensitive to outliers \cite{du2015probability}. Instead of directly finding the nearest neighbor of a source point $ p_{j} $ in the target point cloud, one could consider the local neighborhood of a candidate $ q_{i} $ to reduce the sensitivity of the algorithm to noise.
The point-to-plane ICP variant assumes that the point clouds are locally linear. This local surface can then be defined by its normal vector $\vec{n} $, which is obtained as the smallest eigenvector of the covariance matrix of the points that surround the candidate correspondence $ q_i $. Instead of directly minimizing the Euclidean distance between the corresponding points, we can minimize the scalar projection of this distance on the plane surface defined by the normal vector $ \vec{n}$:
\begin {equation}
\hat {R}, \hat{t} = \argmin_{\hat {R}, \hat {t}} \sum_{i = 1} ^ {N} || ((R  \times p_i + t) - q_j)\times 
\vec{n_i} ||
\end {equation}

\section {3D Key Point Detection and Extraction}
The detection of key points is a very important step in the process of aligning point clouds. Several key point detectors are used to select points of interest in the cloud on which the descriptors are then calculated. The purpose of key point detectors is to determine the points that bear distinctive features in order to obtain good correspondences. Hence the necessity to quantitatively compare the different key point detectors in a common and well-established experimental framework.

\subsection {Evaluation Pipeline}

We run a quantitative evaluation of four key point detectors (extractors). The most important characteristic of a good key point is its \emph{invariance} to common transformations such as rotation, translation, and scale. In order to evaluate the invariance, we extract some key points directly from the source cloud, then we apply a set of transformations to obtain its transformed image (target). Once the key points are extracted from the target, we apply an inverse transformation so that we can compare them with their counterparts in the source. In order for a given detector to be robust against viewpoint change, the applied transformations should not affect the result of key point matching process.

\begin{figure}[th]
	\centering   
	\subfigure[]{\label{Fig_1:a}\includegraphics[height=3.0cm,width=4.0cm]{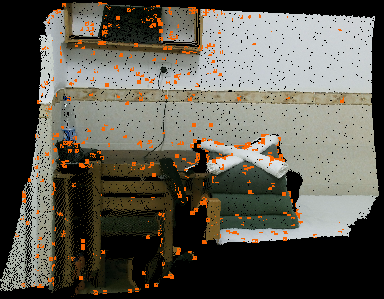}}
	\subfigure[]{\label{Fig_1:b}\includegraphics[height=3.0cm,width=4.0cm]{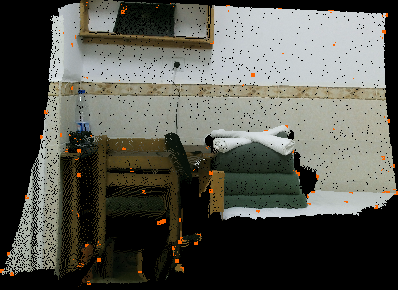}}
	\subfigure[]{\label{Fig_1:c}\includegraphics[height=3.0cm,width=4.0cm]{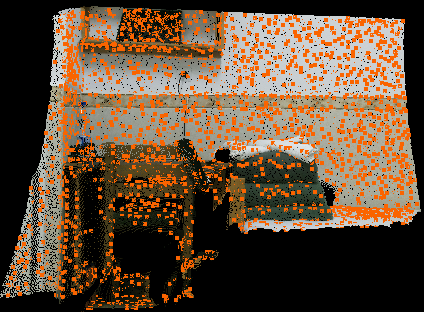}}
	\subfigure[]{\label{Fig_1:d}\includegraphics[height=3.0cm,width=4.0cm]{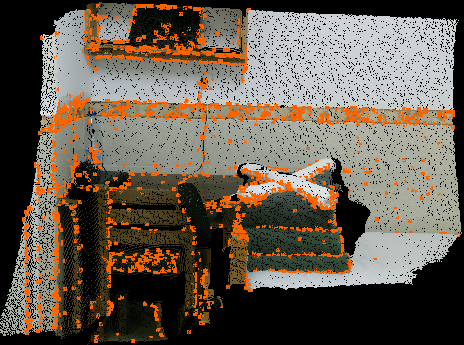}}
	\caption{Key point detectors: (a) SIFT3D, (b) HARRIS3D, (c) ISS3D, (d) SUSAN.}
	\label{Fig_1}
\end{figure}

\subsection {Comparison criterion}
Another important feature of good key point is \emph{repeatability}. This function takes into account the detector's ability to find the same key point in a different view of the same 3D scene. Differences may be due to noise, change of viewpoint, occlusions or a combination of the above. The employed 3D key point repeatability metric was inspired by \cite {tombari2013performance}. A key point extracted from the target view resulting from a rotation, translation or scale is repeatable if the distance $d$ of its closest neighbor, in the source view is less than a given small threshold. Let $ p_ {1} (x_ {1}, y_ {1}, z_ {1}) $ be a key point belonging to the source and $ p_ {2} (x_ {2}, y_ {2}, z_ {2 }) $ its nearest correspondent in the target. The distance $d$ between $ p_ {1} $ and $ p_ {2} $ is calculated by the Euclidean $L_2$ Norm. We evaluate the global repeatability of a detector in relative terms. Given the set of reproducible key points, the relative repeatability is defined as follows:
\begin{eqnarray}
	r=\dfrac{|K|}{|RK|}
\end{eqnarray}
Where $ K $ is the set of all key points extracted from the original view while $ RK $ is the set of repeatable key points.
\subsection {Results}
We test the invariance regarding rotation, translation, and scaling by varying the rotation along the three axes $ x, y, z $. We focused on the analysis of the performance of the algorithms on extreme rotation angles, which can be either small (5$\degree$) or large (40$\degree$). The translation is applied simultaneously to the three axes, and the image displacement distances are obtained randomly. Finally, we test the impact of scaling on key points extractors.
Figure \ref {Fig_1} shows the result of SIFT3D, HARRIS3D, ISS3D and SUSAN detectors on a Kinect point clouds.

\begin{figure}[H]
	\centering   
	\subfigure[]{\label{Fig_2:a}\includegraphics[height=3.0cm,width=4.0cm]{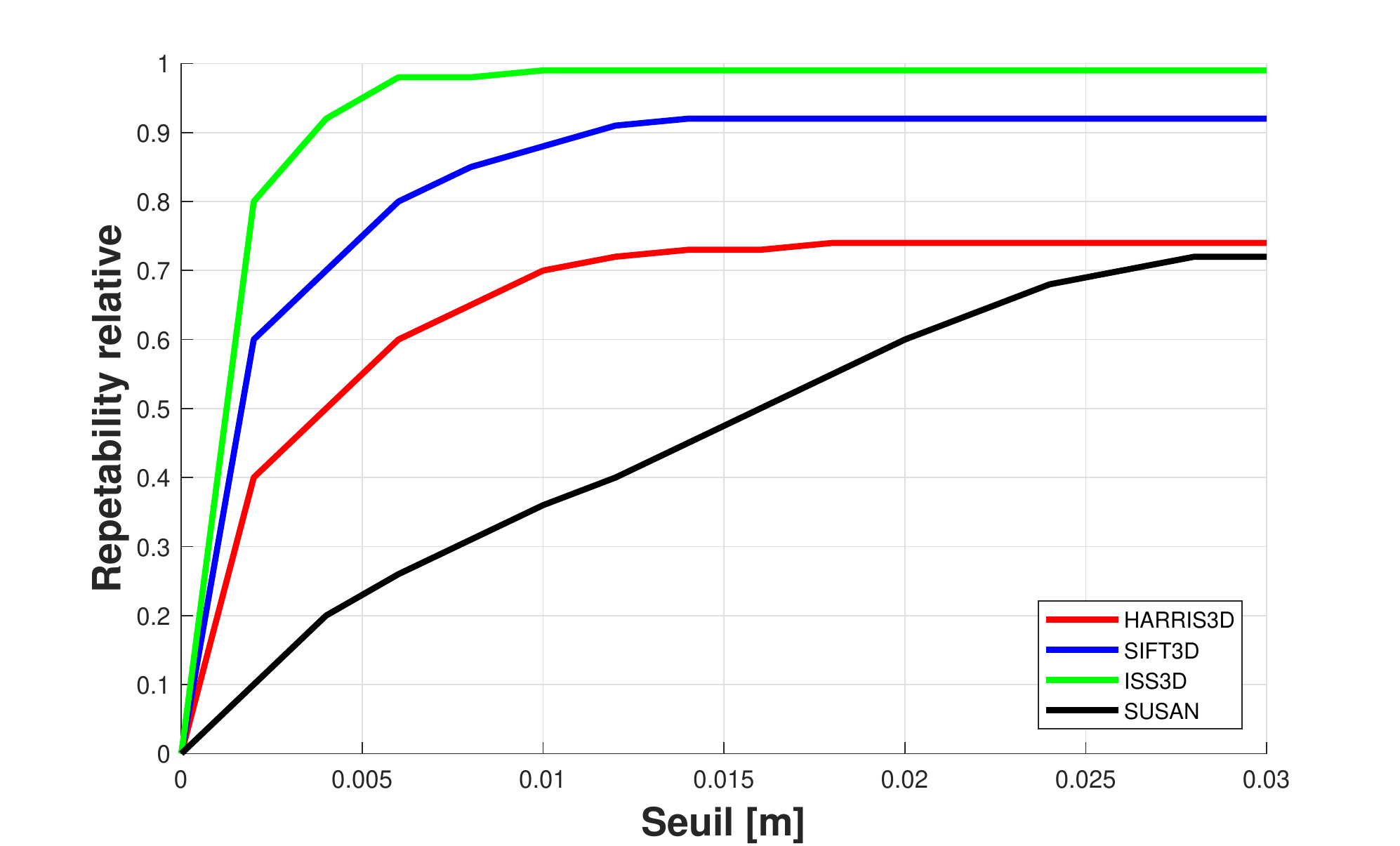}}
	\subfigure[]{\label{Fig_2:b}\includegraphics[height=3.0cm,width=4.0cm]{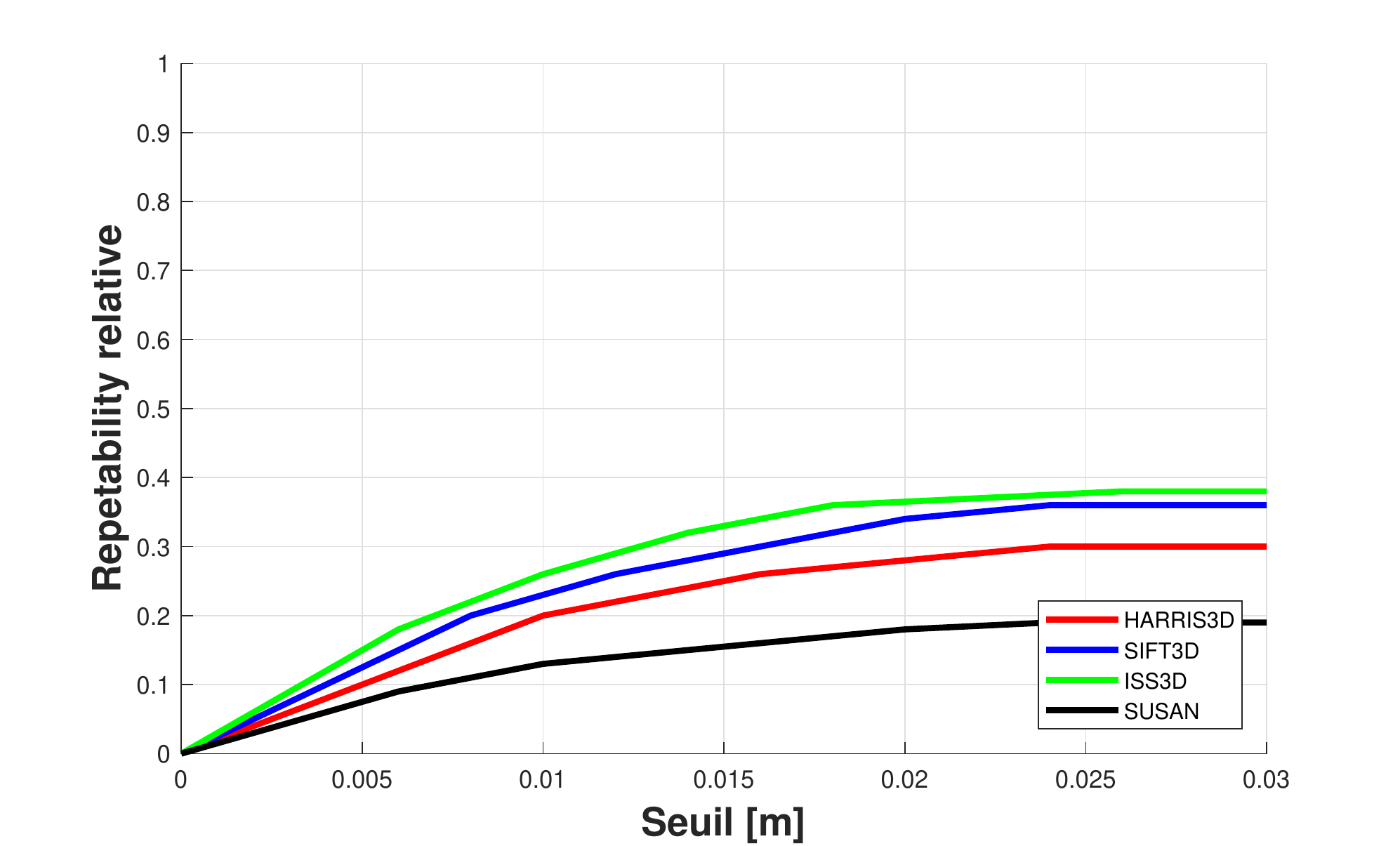}}
	\subfigure[]{\label{Fig_2:c}\includegraphics[height=3.0cm,width=4.0cm]{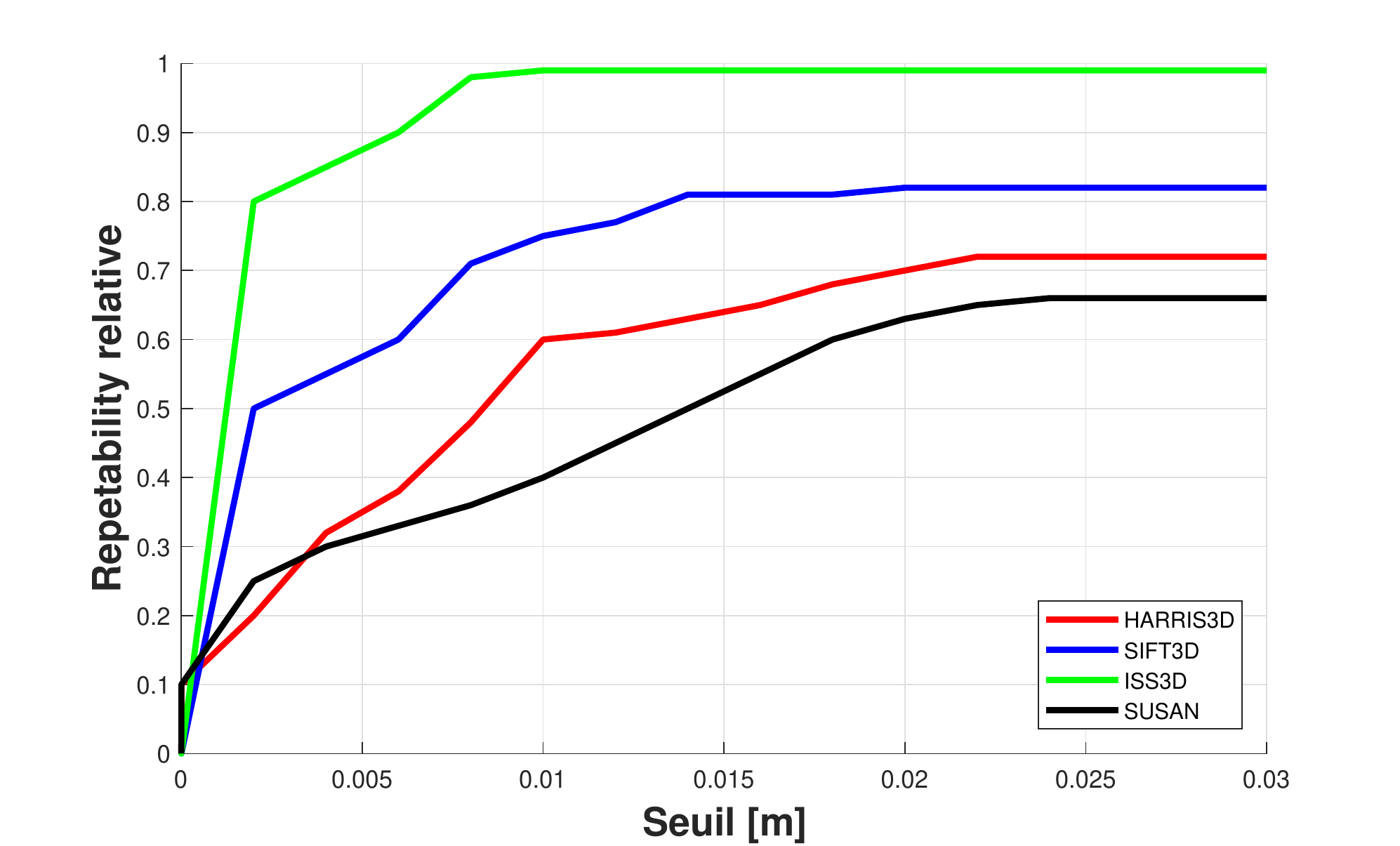}}
	\subfigure[]{\label{Fig_2:d}\includegraphics[height=3.0cm,width=4.0cm]{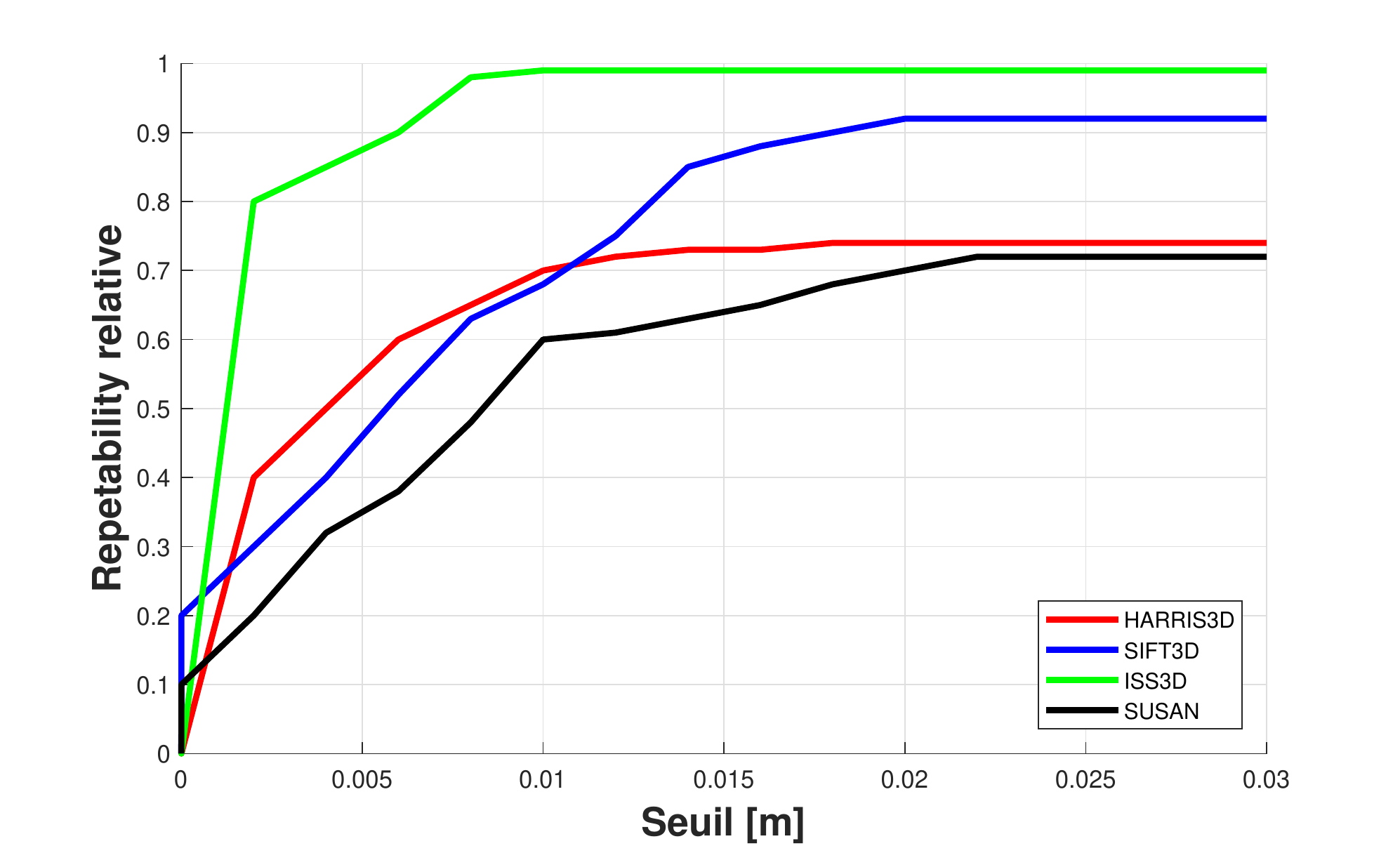}}
	\caption{Relative repeatability. Rotation,(a) 5$\degree$, (b) 40$\degree$, (c) Translation (d) Scaling.}
	\label{Fig_2}
\end{figure}

Figures \ref {Fig_2} shows evaluation results of the different transformations, it is based on the computation of the relative repeatability of the key points between the original point cloud and the transformed one according to a given threshold. In our case, the threshold was set in the interval $[0.0, 3.0] cm$, amounting in 30 equidistant steps. The methods have a relatively large set of parameters to adjust: the values used were those defined by default in PCL library \footnote{http://pointclouds.org/}.

\subsection {Discussion}
Ideally, the results of the method should not change independently of the transformations applied. 
Regarding the repeatability criterion, the detectors have a fairly good performance in general with the exception of SUSAN. For rotation (see Figure \ref {Fig_2}), the increase of the angle tends to proportionally affect the performance of detection.  ISS3D detector provides the best results, however. For example the largest difference between detectors was observed at small angles. In this case, ISS3D delivered almost perfect matching with a error of 0.5 cm. On the other hand, SIFT3D achieves the same performance only for key points at a distance of less than 1.3 cm. For scaling and translation (see Figure \ref {Fig_2}), the methods show very similar results to those obtained for small rotations except for SUSAN, which has a relatively small variance. To conclude, ISS3D and SIFT3D achieve a better invariance than HARRIS3D and SUSAN with respect to the transformations considered in this section.

\section {Key Point Descriptors}
The description of key points is another important step in the alignment process. The quality of key point descriptors has a direct influence on the calculation of the correspondences between point clouds. In addition, the quality of feature matches has a direct impact on the result of alignment. Indeed, a good description is essential for correct alignment. Several key point descriptors have been proposed, and each has advantages and limitations, hence, it is necessary to quantitatively evaluate them in order to determine those that improve the quality of the correspondences. In this section, we will compare FPFH and SHOT descriptors. We chose in the detection step SIFT3D and ISS3D detectors which showed a good invariance to translation and scaling and more efficient than other detectors with respect to rotation. Subsequently, all possible combinations of key point detectors and descriptors will be considered to align two point clouds, in order to find the best combination that minimizes the error.

\subsection {Comparison Criteria : Success rate}
The success rate is calculated after estimating the correspondences based on the descriptors. It is defined by the ratio between the cardinality of the set of correct correspondences and that of all the estimated correspondences. The set of correct matches consists of pairs of key points ($ p_{i} $, $ p_{j} $) that belong to the estimated matches such as for each pair of key points ($ p_{i} $, $ p_{j} $), $ p_{i} $ is actually $ p_{j} $.

\subsection {Comparison Criteria : Misalignment}
The misalignment represents the mean squared error between the original point cloud and the transformed one after alignment. The result is considered of good quality if the error is small.

\subsection {Results}
\begin{figure}[h]
	\centering    
	\subfigure[]{\label{Fig_4:a}\includegraphics[height=3cm,width=4.0cm]{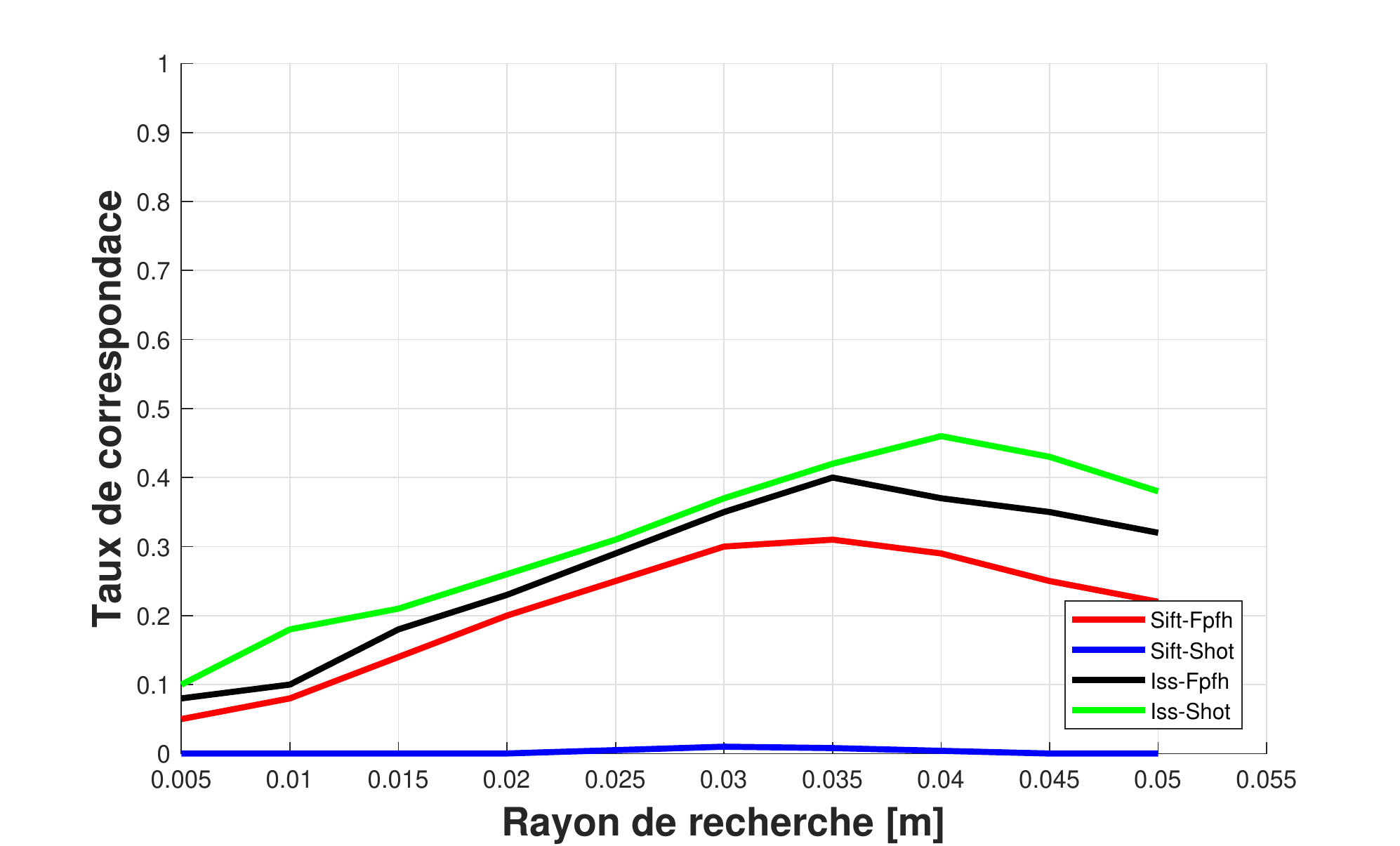}}
	\subfigure[]{\label{Fig_4:b}\includegraphics[height=3cm,width=4.0cm]{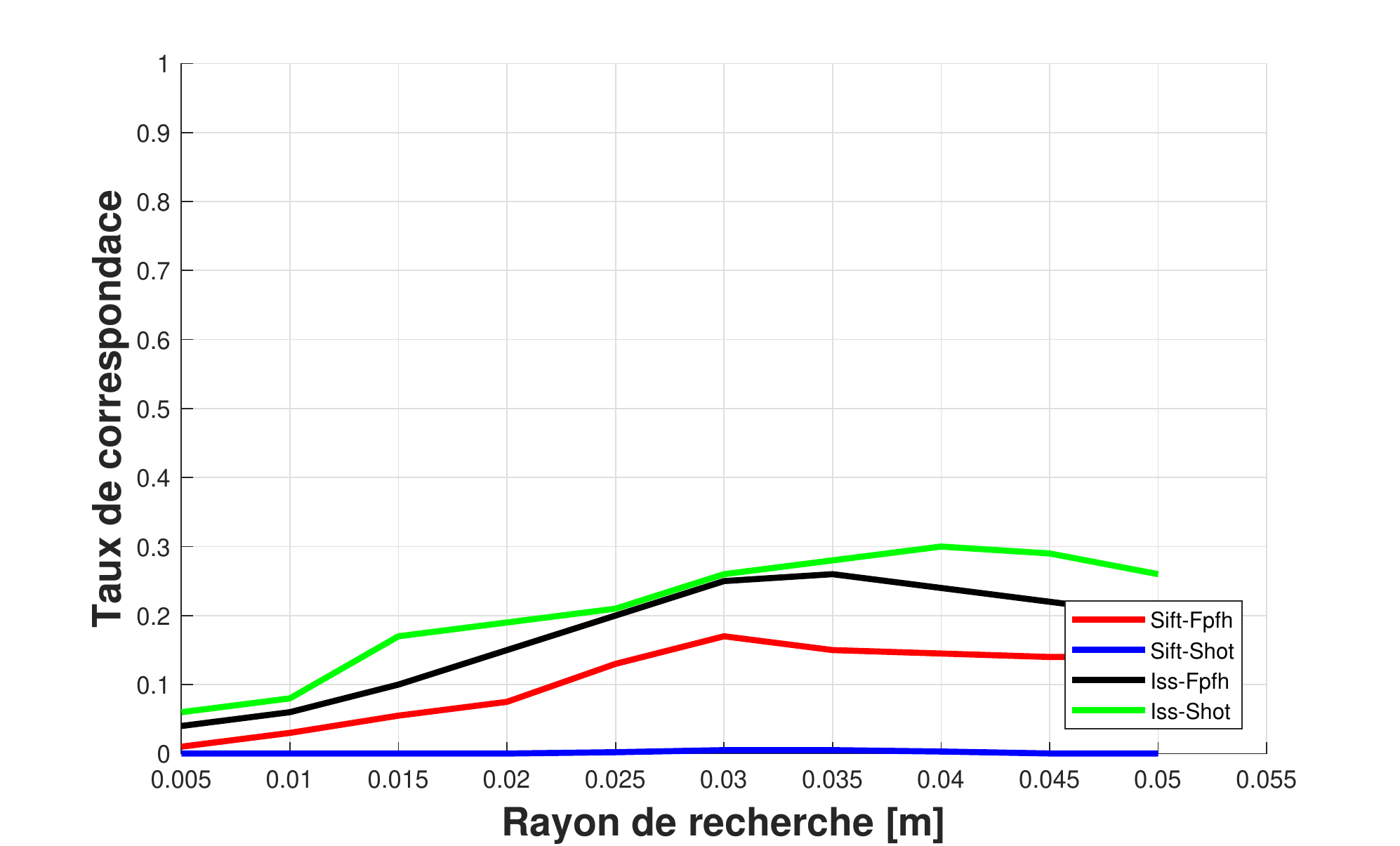}}
	\subfigure[]{\label{Fig_4:c}\includegraphics[height=3cm,width=4.0cm]{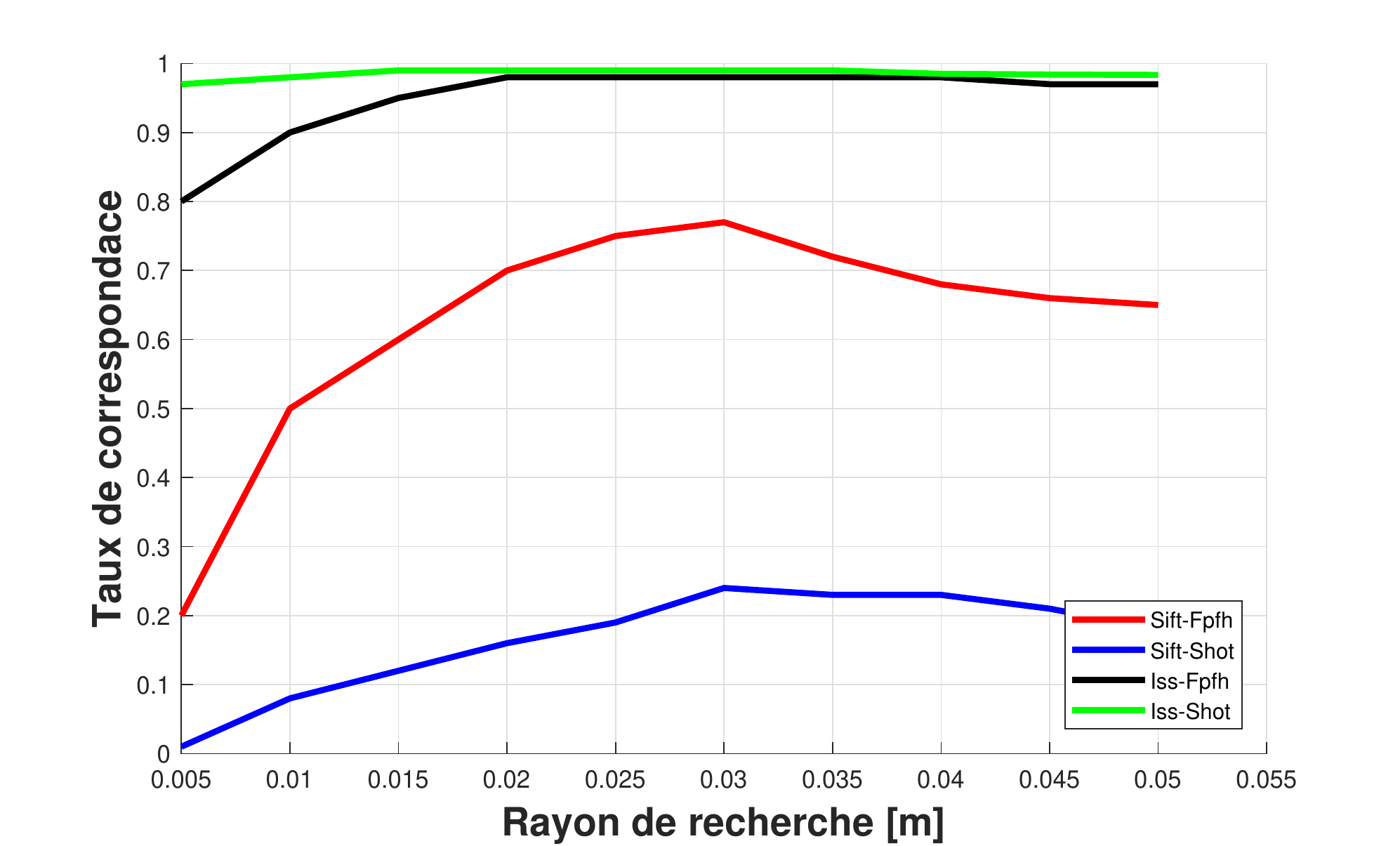}}
	\subfigure[]{\label{Fig_4:d}\includegraphics[height=3cm,width=4.0cm]{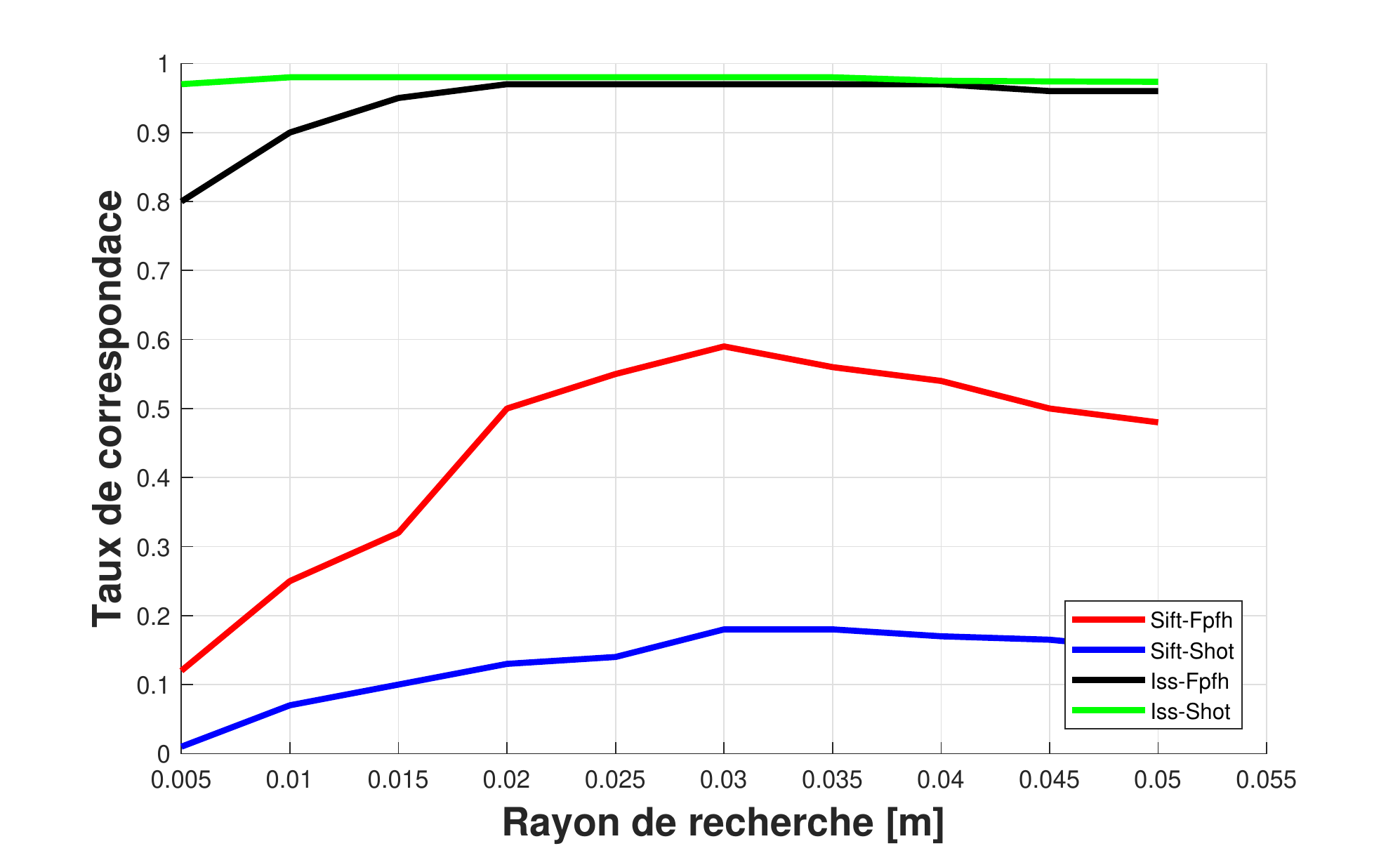}}
	\caption{Success rate after a rotation of: 5$\degree$ (a) and 40$\degree$ (b). Success rate after a translation of: 0.5 $m$ (c) and 1.5 $m$ (d).}
	\label{Fig_4}
\end{figure}

In this section, we study the impact of different combinations of key point detectors and descriptors on the outcome of the matching process and subsequently on the quality of alignment to evaluate their invariance to common 3D transformations. We run the alignment pipeline on the source and target point clouds. All detector combinations (SIFT3D and ISS3D) and descriptors (FPFH and SHOT) are tested. ICP algorithm with its point-to-point and point-to-plane variants are used in the refinement step.
Regarding the first evaluation part, which is based on the calculation of matching success rate for different combinations of detectors and descriptors, Figure \ref {Fig_4} shows the success rate, here, descriptor search radius varies between $[0.0, 5.0] cm$ with a step of 0.005 cm.
The second evaluation step is based on the calculation of alignment error for various combinations of detectors and descriptors, Figure \ref {Fig_6} shows the relationship between the errorand search radius.

\begin{figure}[h]
	\centering     
	\subfigure[]{\label{Fig_5:a}\includegraphics[height=3cm,width=4.0cm]{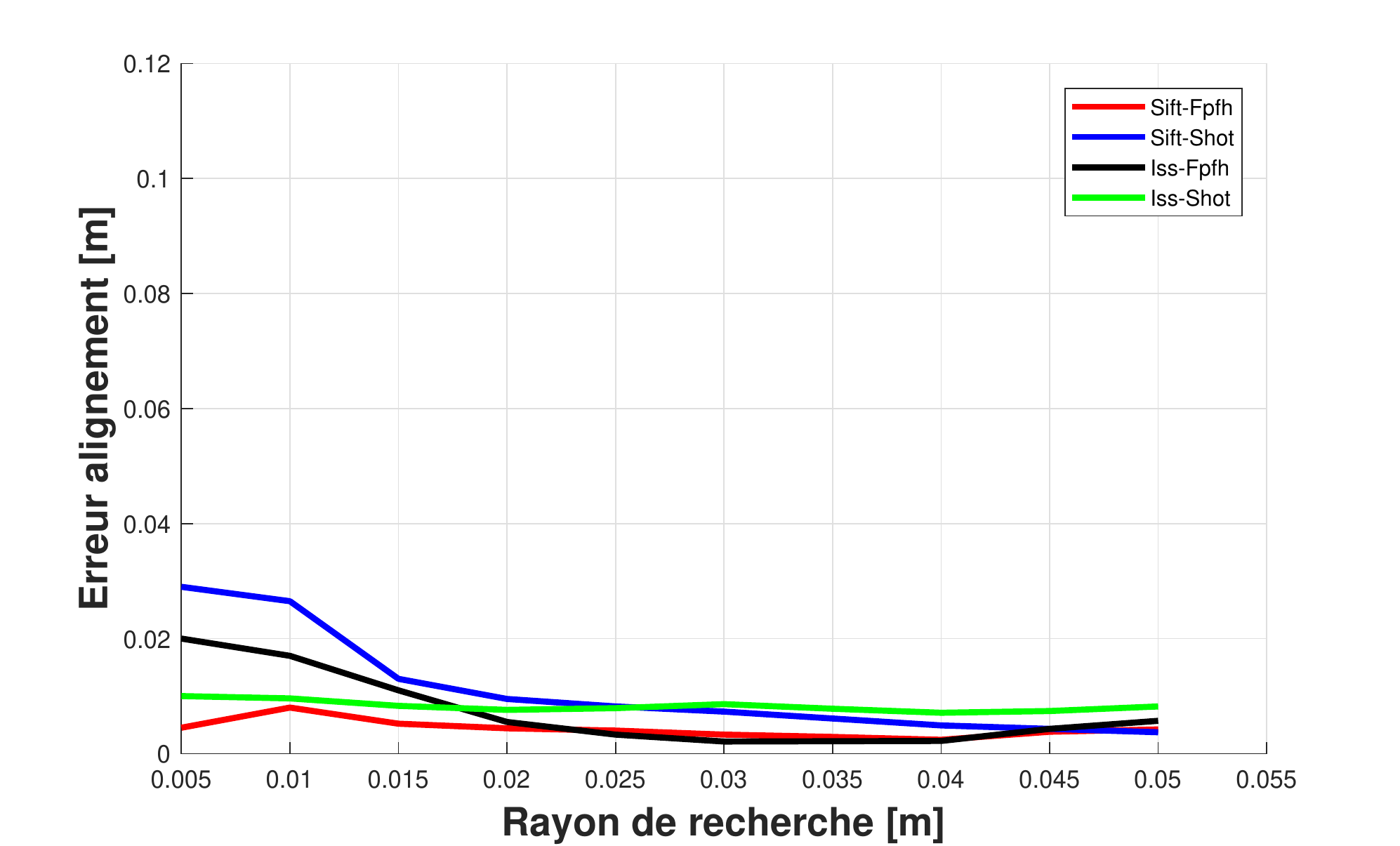}}
	\subfigure[]{\label{Fig_5:b}\includegraphics[height=3cm,width=4.0cm]{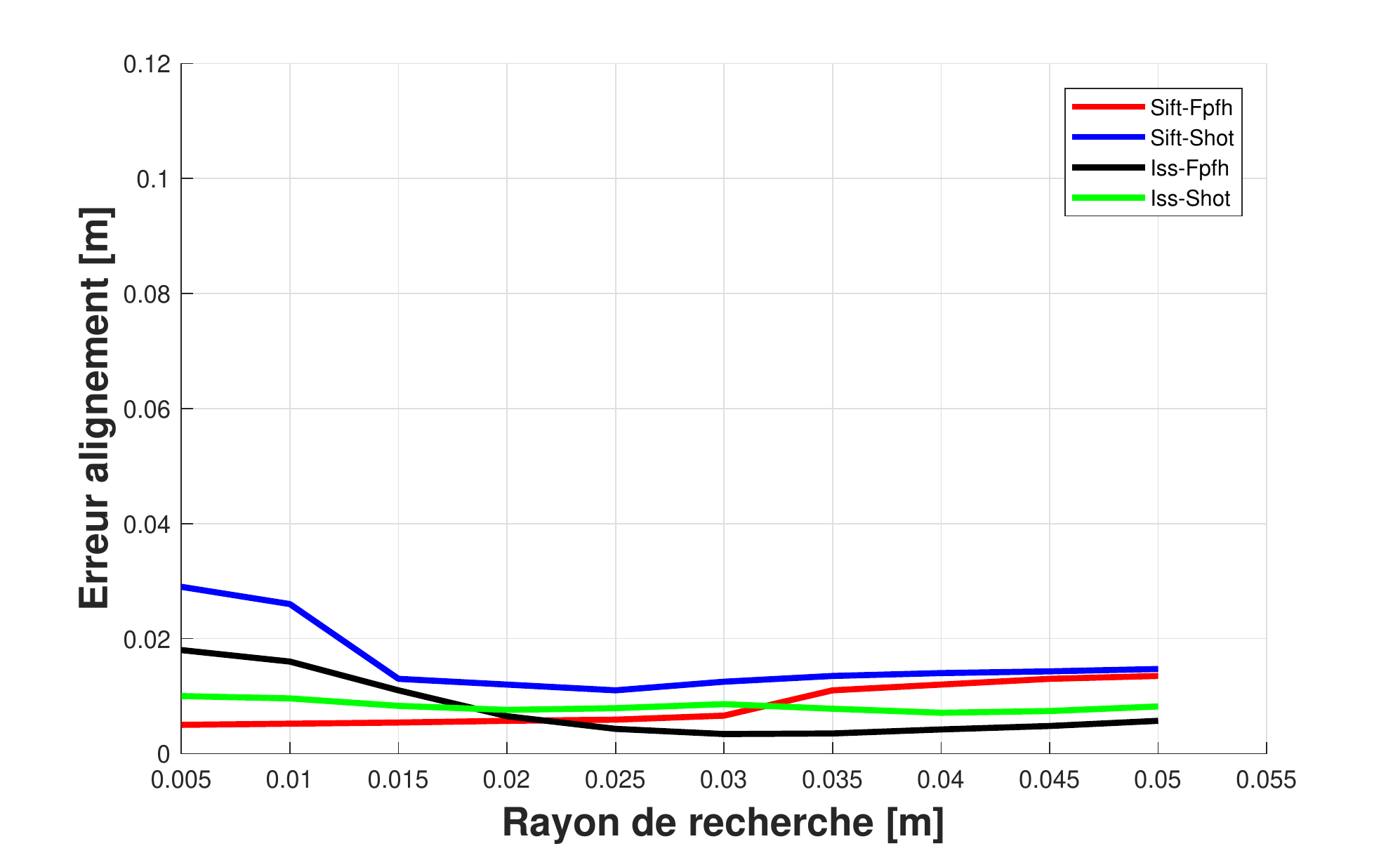}}
	\subfigure[]{\label{Fig_5:c}\includegraphics[height=3cm,width=4.0cm]{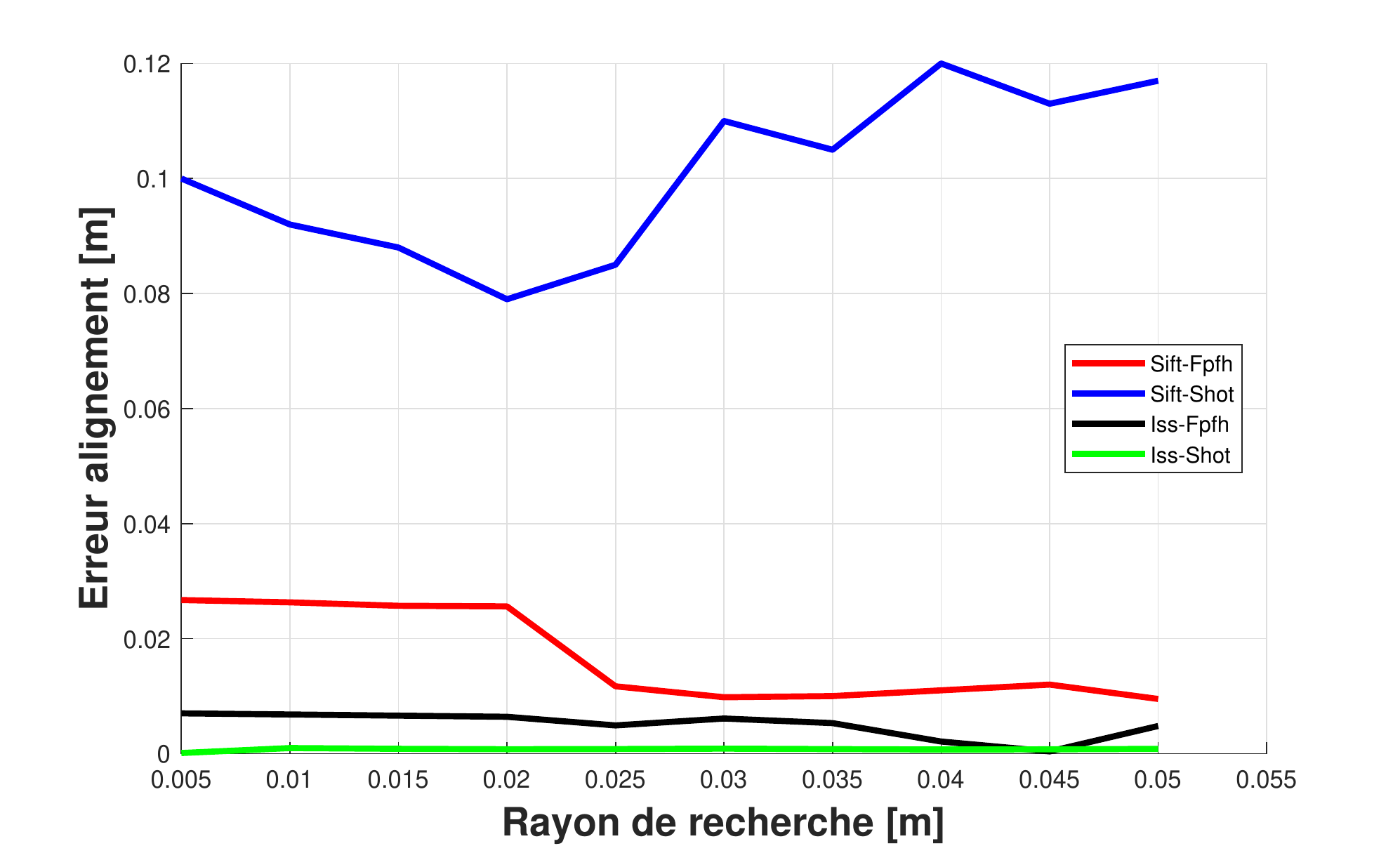}}
	\subfigure[]{\label{Fig_5:d}\includegraphics[height=3cm,width=4.0cm]{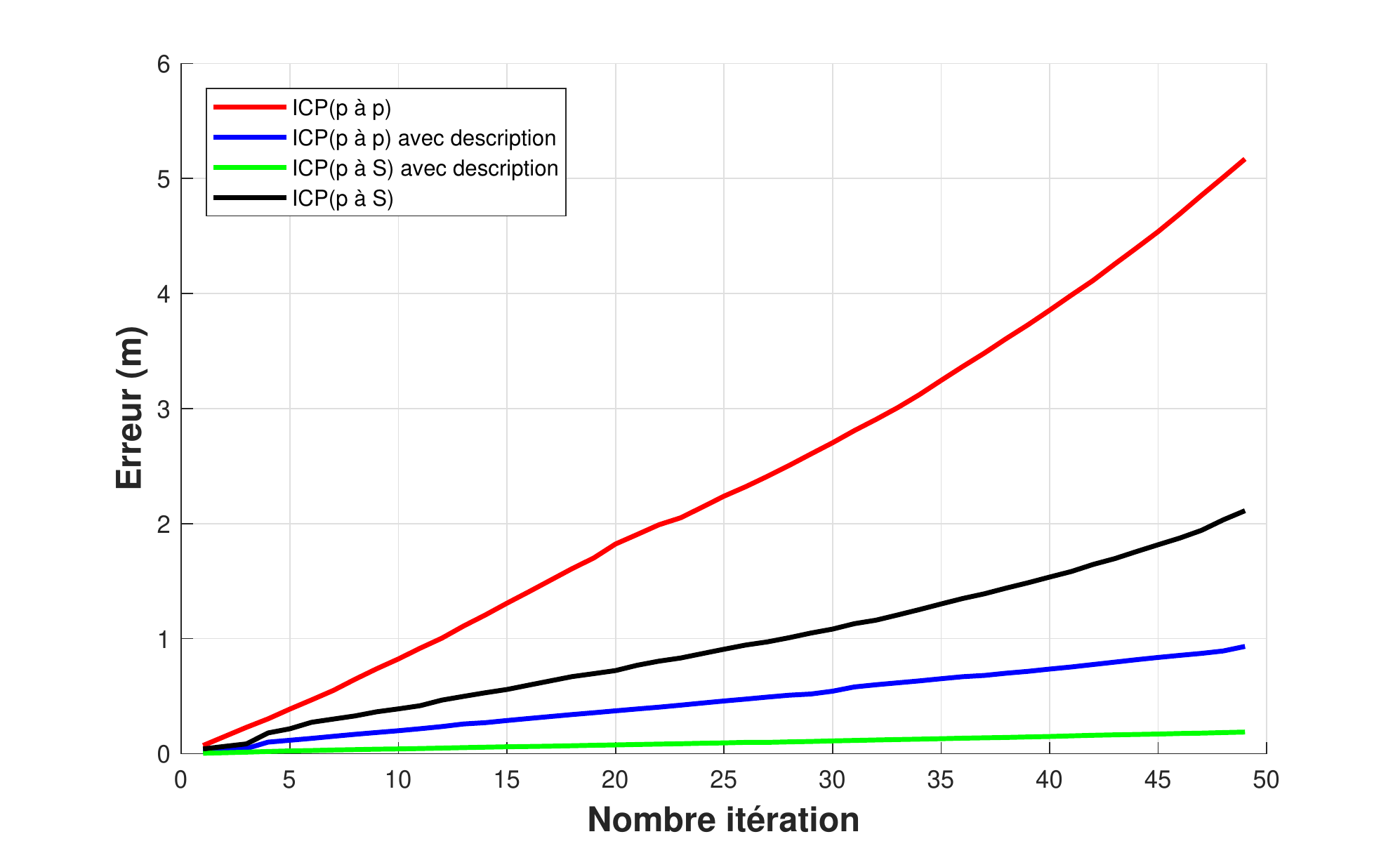}}
	\caption{Misalignment after a rotation of: (a) 5$\degree$, (b) 40$\degree$, (c) Misalignment after a translation of: $ 1 m $. (d) Variation of the cumulative error during the refinement process.}
	\label{Fig_5}
\end{figure}

\subsection {Discussion}
We start with the evaluation of matching success rates with different combinations of detectors and descriptors. Such combinations usually show good performance except for SIFT3D-SHOT. As for the rotation (see Figure \ref {Fig_4:a}, \ref {Fig_4:b}), these combinations have an average performance; the increase of the angle of rotation worsens the results. The combinations ISS3D-SHOT and ISS3D-FPFH provide the best results for rotation. They achieve their maximum performance for a search radius in the interval [3.5, 4.5] cm. Regarding translation (see Figure \ref {Fig_4:c}), the use of both ISS3D-SHOT and ISS3D-FPFH leads to a success rate greater than 0.99 for ISS3D-SHOT and greater than 0.9 for ISS3D-FPFH, so these two combinations are invariant to translation. The conclusion we can draw from the evaluation is that the use of ISS3D detector offers, for both combinations, a higher success rate than SIFT3D detector.

In the second phase of the discussion, we test the misalignment with different combinations of detectors and descriptors as a function of search radius, these combinations present close results for a rotation of 5$\degree$ with an error less than 1 cm. As the rotation angle increases, the misalignment increases proportionally with a relatively higher rate for the SIFT3D-SHOT and SIFT3D-FPFH combinations than the other two combinations. With the translation, the use of the ISS3D-SHOT and ISS3D-FPFH combinations leads to a good alignment with almost null error, SIFT3D-FPFH leads to a good alignment with an average error of 2.0 cm, whereas the combination SIFT3D-SHOT leads to misalignment with an average error of 10.0 cm.

The first observation we have drawn from the previous evaluation is that the alignment result is invariant to rotation, which is contradictory to the result of the matching process. This implies that the refinement step performed by the ICP is invariant to rotation. The second finding we made is that the alignment result is invariant to translation only for combinations that were already invariant to translation in the matching step. Therefore, the result of refinement depends directly on the matching process for the translation. Overall, ISS3D-SHOT and ISS3D-FPFH combinations outperform SIFT3D-SHOT and SIFT3D-FPFH with respect to the invariance to the transformations.

\section {Refinement}
The final part of the alignment process is the refinement of the output of sparse alignment. The most commonly used algorithm is ICP (Iterative Closest Point) \cite{du2015probability}. This algorithm has become a standard in computer vision because of its reliability and flexibility. Given an initial coarse alignment, ICP associates the points of two different views according to the nearest neighbor criterion, it then proceeds through the minimization of the mean squared distance to obtain the transformation. Two widely used variants of the algorithms are point-to-point and point-to-plane.
We test ICP’s ability to make a complete reconstruction of a 3D scene from several images taken at different parts of the scene. The two variants are tested with and without key point descriptors, after which a comparison is made to determine the more performant variant. For instance, the variant that uses a coarse alignment result was obtained with the combination ISS3D-SHOT, which has the best performance compared to its competitors.
\subsection {Comparison criterion}
Generally, 3D point cloud alignment techniques introduce a non-negligible cumulative error, when several point clouds are aligned. Such an error leads to an erroneous result, hence, we use it as a comparison criterion.
\subsection {Test}
\begin{figure}[h]
	\centering 
	\includegraphics[width=\linewidth]{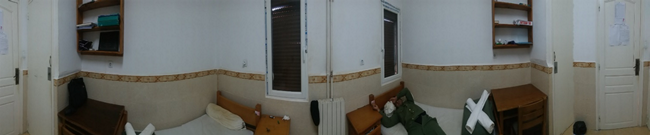}
	\caption{360-degree picture of the scene}
	\label{Fig_6}
\end{figure}

We tested the two variants on several images taken, at different viewpoints in the real scene. The main aim of the test is to check the reliability of the algorithm in the complete scene reconstruction task. Figure \ref {Fig_6} shows a 360-degree picture of the subject indoor scene.

In order for the alignment to work properly, image acquisition is done in such a way that each pair of successive views overlap with each other.

\subsection {Results}
Regarding the evaluation of the whole reconstruction of the scene, which is based on the calculation of the cumulative error generated by ICP, Figure \ref {Fig_5:d} shows the evolution of the latter over the intervening samples.

\subsection {Discussion}
As the first impression from Figure \ref {Fig_5:d}, the use of the ICP algorithm after the initial alignment based on the calculation of key point descriptors gives a better result than without descriptors. In addition, the result obtained by the point-to-plane ICP variant is better than the point-to-point one.

The graph in Figure \ref{Fig_5:d}, which represents the progressive variation of the cumulative error during the reconstruction of the scene, confirms the visual result. Using the point-to-plane variant after initial alignment gives the best result with an overall error of 18cm, whereas the error of point-to-point variant is 98cm. The application of ICP without initial alignment performs poorly as it results in an error of 2.14m for the point-to-plane variant and an error of 5.17m for the point-to-point.

The most important conclusion from these results is that the use of the initial alignment based on the calculation of key point descriptors helps significantly in the refinement task as it allows to avoid falling in local minima; in addition, the point-to-plane variant is more efficient.

\section{Conclusion and future works}
In this work, we provided a comparative study for each stage of a typical coarse-to-fine registration procedure. We began by making an experimental comparison between 3D key point detectors to test their invariance to different transformations (rotation, scaling, and translation). 

Overall, SIFT3D and ISS3D gave the best results in terms of repeatability, and ISS3D proved to be the most invariant. Then we made a comparison between different combinations of key point detectors and descriptors for point cloud registration. Overall, ISS3D-SHOT and ISS3D-FPFH combinations gave the best results. We ended up making a comparison between ICP refinement algorithm variants to figure the one that allows a complete reconstruction of a 3D scene with less error. The point-to-plane variant showed superiority in performance.

To sum up, the most efficient pipeline configuration from a reconstruction quality point of view is that of ISS3D detector and SHOT descriptor in the feature-based coarse alignment phase and ICP point-to-plane dense alignment. Our evaluation results show that this configuration can produce the best 3D reconstruction quality for an indoor scene scanned with Kinect V2.

As future works, we noticed in the correspondence estimation process that most of the tested descriptors are sensitive to rotation. We will try to identify the cause of the problem and to tackle it with more robust alignment strategies. We plan to try alternative 3D data acquisition devices to cope with noise inherent in sensing systems.

\printbibliography
\end{document}